\documentclass[conference]{IEEEtran}
\IEEEoverridecommandlockouts

\usepackage{cite}
\usepackage{amsmath,amssymb,amsfonts}
\usepackage{algorithmic}
\usepackage{graphicx}
\usepackage{textcomp}
\usepackage{xcolor}
\def\BibTeX{{\rm B\kern-.05em{\sc i\kern-.025em b}\kern-.08em
    T\kern-.1667em\lower.7ex\hbox{E}\kern-.125emX}}
\usepackage{graphicx}
\graphicspath{{figures/}}
\usepackage{hyperref} 

\begin{document}

\title{Equity-Aware Geospatial AI for Forecasting Demand-Driven Hospital Locations in Germany}

\author{
\begin{tabular}{ccc}
\\
\textbf{Piyush Pant}$^{*}$ & \textbf{Marcellius William Suntoro}$^{*}$ & \textbf{Ayesha Siddiqua}$^{*}$ \\
\small \texttt{piyushpant15@gmail.com} & 
\small \texttt{masu00001@stud.uni-saarland.de} & 
\small \texttt{aysi00001@stud.uni-saarland.de}
\end{tabular}
\\[1.2em]
\begin{tabular}{cc}
\textbf{Muhammad Shehryaar Sharif}$^{*}$ & \textbf{Daniyal Ahmed}$^{*}$ \\
\small \texttt{mush00003@stud.uni-saarland.de} & 
\small \texttt{dakh00001@stud.uni-saarland.de}
\end{tabular}
\\[1em]

\textit{Department of Computer Science, Saarland University, Germany} \\
\\
\textbf{$^{*}$All authors contributed equally}\\[0.3em]
}

\maketitle

\begin{abstract}
This paper presents EA-GeoAI, an integrated framework for demand forecasting and equitable hospital planning in Germany through 2030. We combine district-level demographic shifts, aging population density, and infrastructure balances into a unified Equity Index. An interpretable Agentic AI optimizer then allocates beds and identifies new facility sites to minimize unmet need under budget and travel-time constraints. This approach bridges GeoAI, long-term forecasting, and equity measurement to deliver actionable recommendations for policymakers.

\end{abstract}

\vspace{1em} 
{\scriptsize
\noindent\textbf{Application:} \url{https://equity-aware-geospatial-ai-project.streamlit.app/} \\
\textbf{Codebase:} \url{https://github.com/mwsyow/equity-aware-geospatial-ai-project/}
}
\vspace{1em} 

\section{Introduction}

Access to hospital care close to home can mean the difference between life and death, yet even in well-resourced regions like Saarland, some communities face longer waits and greater travel burdens than others~\cite{luo2003,timm2020,stellmacher2022}. As Saarland’s population ages, these gaps in access risk widening, with under-served districts bearing the greatest health and social costs~\cite{helmig2021,verma2022}. Our project addresses this problem by developing a unified pipeline that combines spatial data, demographic forecasting, and equity-aware metrics to guide the placement of hospital beds where they are needed most~\cite{bohm2021,meyer2023}.

Existing studies have explored elements of this challenge, such as mapping travel times~\cite{luo2003,stellmacher2022}, projecting population changes~\cite{verma2022,helmig2021}, or measuring socioeconomic deprivation~\cite{bohm2021,meyer2023}—but they stop short of linking these pieces into a single decision tool. We fill that gap by integrating municipal boundaries from Destatis, district-level elderly population forecasts, and detailed travel-time matrices into two core indices: an Equity Index that weights unmet demand by vulnerability and an accessibility index that captures the ease of reaching care~\cite{daskin2011,rauch2019}. Building on these indices, we introduce a simple planning agent that iteratively allocates optimal locations for new hospitals, ensuring that each placement maximally improves both fairness and efficiency~\cite{crooks2013,yin2021,zhang2024}.

Recent advances in AI have shown that principled optimization and evaluation can guide models toward socially beneficial outcomes~\cite{pant2025sft}, highlighting the importance of interpretable and responsible AI in decision-making. In particular, methods for aligning AI systems to desired objectives demonstrate how optimization under constraints can improve decision quality, which conceptually motivates our interpretable agentic AI framework for equitable hospital planning.

To validate our approach, we compare it against the status quo of existing hospitals and a population-weighted heuristic. We also run ablation studies focused solely on demand, deprivation, or travel time. Evaluation uses six metrics—equity score, mean and median travel time, P95 travel time, hospital-demand ratio, over-served count, and a Lorenz-curve-based Gini coefficient—to provide a comprehensive view of performance.

\subsection*{Organization}

Section II reviews related work, highlighting existing gaps in the literature and how the proposed approach addresses them. Section III outlines the project objectives, research design, and a comprehensive description of the modeling pipeline. Section IV details the experimental methodology, including implementation setup and the performance metrics employed to assess model effectiveness. Section V presents the experimental results, accompanied by a critical discussion of the findings and their limitations. Finally, Section VI concludes the study by summarizing the key contributions and outlining potential directions for future research.

\section{Related Work}

Health geography has long relied on the Two‐Step Floating Catchment Area (2SFCA) method to quantify spatial access to care.  First introduced by Luo and Wang in 2003, 2SFCA remains a foundational tool for measuring how easily people can reach health services \cite{luo2003}.  In Germany, nationwide 2SFCA maps have been produced for stroke units \cite{timm2020} and obstetric clinics \cite{stellmacher2022}, highlighting areas with poor geographic access.

Existing studies have applied machine learning to forecast hospital demand.  Helmig \emph{et al.} used gradient boosting to predict surgical caseloads at the NUTS‐3 level, achieving a 15 percent reduction in error compared to naïve trend models \cite{helmig2021}.  Verma \emph{et al.} incorporated age‐band projections and seasonality adjustments in the U.K., demonstrating how demographic shifts affect short‐term demand \cite{verma2022}.  These approaches advance forecasting but do not translate predictions directly into placement decisions. Also, approaches from AI alignment research~\cite{pant2025sft} demonstrate how optimization under constraints can improve model behavior, inspiring our interpretable agentic AI framework for equitable hospital planning.

Socioeconomic deprivation analyses have overlaid the German Index of Socioeconomic Deprivation (GISD) on accessibility scores to reveal equity gaps \cite{bohm2021, meyer2023}.  While these studies expose critical mismatches between need and access, they stop short of prescribing optimized solutions.

Facility location models traditionally use mixed‐integer linear programming to minimize travel under capacity and budget constraints \cite{daskin2011}.  Rauch introduced a hybrid genetic‐algorithm/MILP approach for emergency department coverage in Germany \cite{rauch2019}.  Chen \emph{et al.} weighted facility siting by social vulnerability in China, underscoring the importance of equity \cite{chen2021}.

Geospatial AI brings new methods to healthcare planning.  Crooks and Wise simulated ambulance dispatch with agent‐based models \cite{crooks2013}.  Yin applied convolutional neural networks to predict service catchments \cite{yin2021}.  Zhang deployed reinforcement learning for field hospital placement in disaster relief scenarios \cite{zhang2024}.  These works showcase AI’s potential but do not integrate forecasting, equity metrics, and optimization in a unified pipeline.

\textbf{Summary.}  In sum, four strands—spatial accessibility, demand forecasting, deprivation mapping, and optimization or GeoAI—have advanced the field, yet remain siloed.  Our Equity‐Aware Geospatial AI Pipeline bridges these areas to deliver an end‐to‐end framework for equitable hospital planning.

\section{Methodology}

\subsection*{Project Objective}
The primary objective of this study is to develop and evaluate an equity-aware, data-driven framework for optimizing hospital locations within the Saarland region. The project aims to address disparities in healthcare accessibility by leveraging geospatial analysis and AI-based modeling to propose hospital placements that minimize travel time and improve service equity across districts. By integrating demographic, epidemiological, and infrastructural data, our approach seeks to advance the current understanding of spatial healthcare equity and provide actionable insights for policy makers.

\subsection*{Research Design}
This research adopts a quantitative, computational approach, combining geospatial data analysis with machine learning and optimization techniques. The study is observational in nature, utilizing publicly available datasets on hospital locations, population demographics, disease incidence, and travel times. The workflow is modular, with separate components for data preprocessing, index computation, demand forecasting, and model evaluation. The design ensures reproducibility and extensibility, allowing for the integration of additional data sources or alternative modeling strategies as needed.

\subsection*{Data Analysis}
Data analysis is conducted through a series of Python modules that process and synthesize information from multiple sources. Key steps include:
\begin{itemize}
    \item \textit{Data Preprocessing:} Raw data on hospital locations, district populations, and disease incidence are cleaned and harmonized. Geospatial coordinates are standardized, and travel time matrices are computed using network analysis.
    \item \textit{Index Computation:} Several indices are calculated to quantify healthcare demand, deprivation, elderly share, hospital capacity, and accessibility. For example, the demand forecast index leverages ARIMA time series models to predict future healthcare needs at the district level.
    \item \textit{Equity and Accessibility Metrics:} Metrics such as the equity index, accessibility score, and hospital facility demand ratio (HFDR) are computed to evaluate the spatial distribution of healthcare resources and identify underserved areas.
    \item \textit{Evaluation:} The effectiveness of different hospital location models is assessed using these metrics, with results visualized through interactive dashboards and geospatial maps.
\end{itemize}

\subsection*{Model Development}

\begin{figure}[ht]
    \centering
    \includegraphics[width=0.8\linewidth]{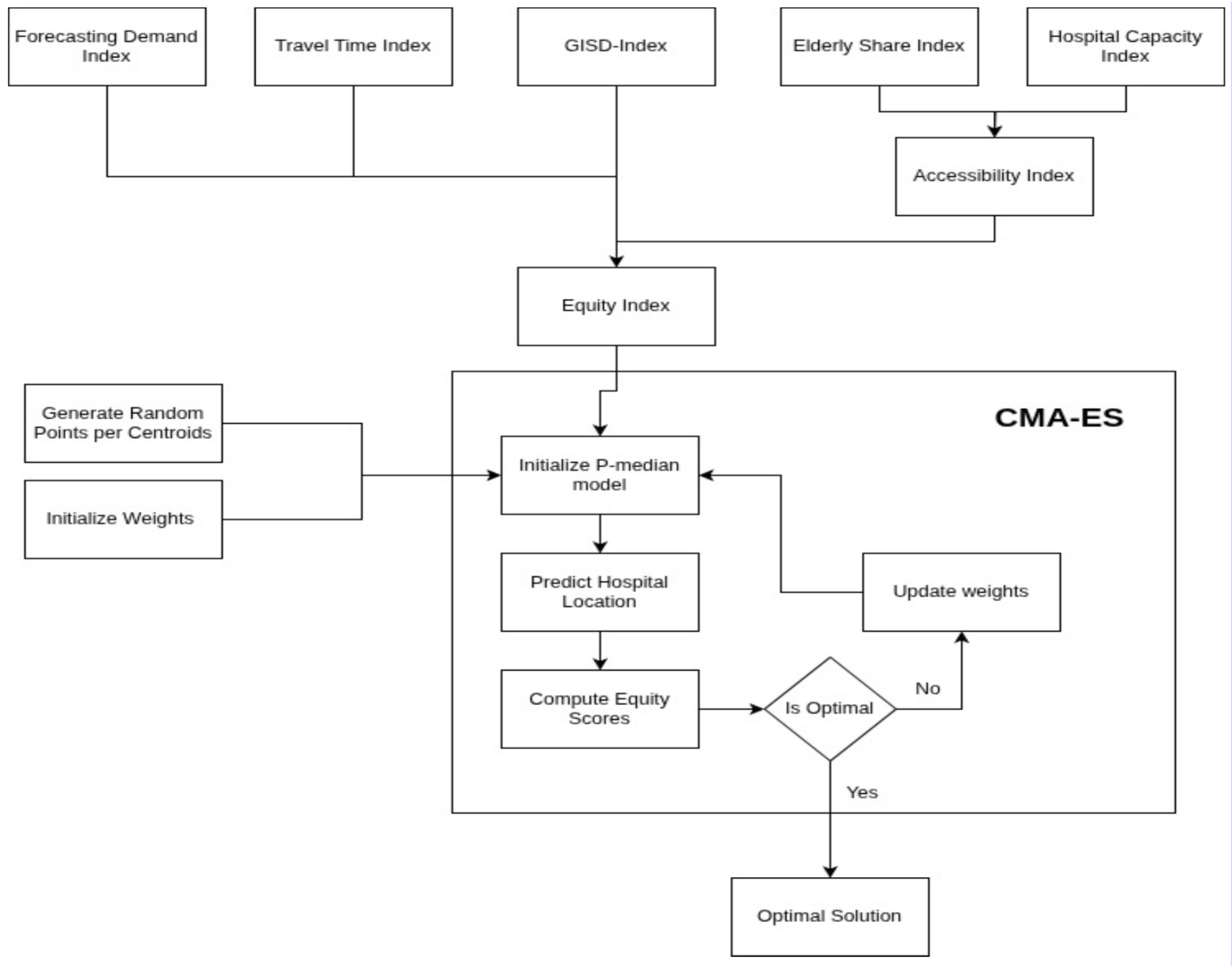}
    
    \caption{Agentic Model Pipeline}
    \label{fig:model_diagram}
\end{figure}
The core of the project is an agentic AI planner that formulates the hospital location optimization as a mathematical programming problem. The model incorporates the following components:
\begin{itemize}
    \item \textit{Objective Function:} Minimize the total or average travel time from population centroids to the nearest hospital, subject to equity constraints.
    \item \textit{Constraints:} Include capacity limits, minimum service requirements, and spatial equity thresholds to ensure fair access across all districts
    \item \textit{Components:} The model integrates outputs from demand forecasting, deprivation indices, and accessibility calculations. It is implemented using Python and leverages libraries such as \texttt{networkx} for graph-based analysis and \texttt{statsmodels} for time series forecasting.
    
\end{itemize}
Figure 1 shows how the proposed method works. Our pipeline begins by calculating a district-level Equity Index. We then generate candidate hospital locations near population-dense centroids within each district. A linear programming model is used to determine the optimal placement of new hospitals, with an objective function that simultaneously maximizes the Equity Index and minimizes travel time between hospitals and their assigned centroids. Constraints ensure that hospitals are only assigned to centroids reachable within 30 minutes and are not placed too close to existing or other proposed hospitals. To optimize the objective function's weighting parameters, we employ the Covariance Matrix Adaptation Evolution Strategy (CMA-ES), using the average travel time from assigned hospitals to centroids as the loss function.

\subsection*{Agentic Nature of the Optimization Framework}
The project exhibits agentic characteristics by autonomously integrating, analyzing, and acting upon diverse geospatial and demographic data to optimize hospital locations. At its core, the system functions as an intelligent agent: it perceives the healthcare landscape through multiple data sources (including population, demand, deprivation, and accessibility indices), reasons about the current and projected needs of the region, and proactively generates optimized solutions that balance efficiency and equity. The agentic AI planner not only evaluates existing conditions but also simulates and recommends new hospital placements, adapting its strategy based on constraints and objectives defined by stakeholders. This agentic approach enables the system to move beyond static analysis, actively shaping outcomes in a goal-directed manner and providing actionable insights for decision-makers in healthcare planning.

\section{Experiments}

\subsection*{Data Collection}
The experimental phase of this project is grounded in the integration and analysis of diverse, high-quality datasets relevant to healthcare accessibility and hospital location optimization in the Saarland region. Data collection was conducted through the aggregation of multiple publicly available and institutional sources, ensuring both breadth and depth in the information utilized. The primary data sources include:

\begin{itemize}
    \item \textbf{Hospital Locations and Capacities:} Data on existing hospitals, including their geographic coordinates, bed capacities, and service types, were obtained from the official Krankenhausverzeichnis (Hospital Directory) and regional health authority records.
    \item \textbf{Population Demographics:} District-level population statistics, age group distributions, and projections were sourced from governmental census data and statistical offices. Special attention was given to the elderly population, as this demographic is particularly relevant for healthcare planning.
    \item \textbf{Disease Incidence and Hospital Inpatients:} Historical records of disease incidence and hospital inpatient numbers were collected to inform demand forecasting and to capture temporal trends in healthcare utilization.
    \item \textbf{Geospatial Data:} Shapefiles and geospatial boundaries for districts and municipalities were incorporated to enable spatial analysis and mapping. These were sourced from official GIS repositories.
    \item \textbf{Travel Time Matrices:} Using the geographic coordinates of hospitals and population centroids, travel time matrices were computed via network analysis, leveraging road network data and routing algorithms.
\end{itemize}

\begin{figure}[ht]
    \centering
    \includegraphics[width=0.8\linewidth]{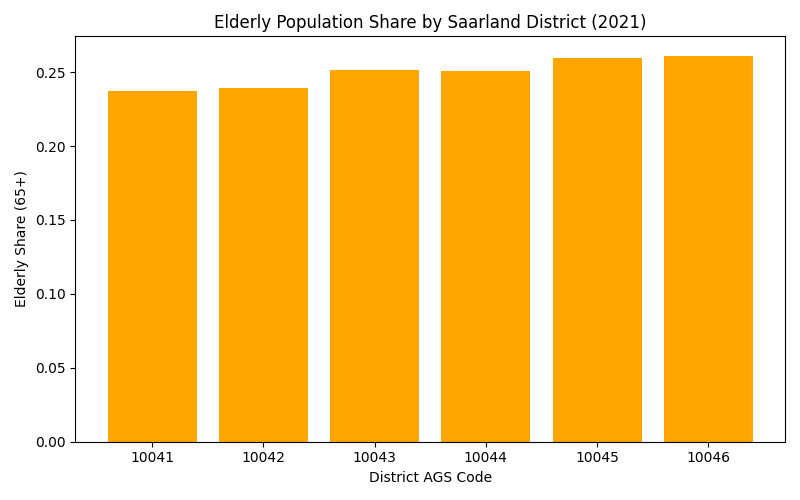}
    \caption{Elderly population share (age 65+) by Saarland district in 2021}
    \label{fig:elderly_share}
\end{figure}

The data collection process emphasized data quality, completeness, and consistency. Where necessary, data cleaning and harmonization procedures were applied to resolve discrepancies in formats, units, and coordinate systems. No primary data collection (e.g., surveys or interviews) was conducted; instead, the study leverages the richness of existing datasets. Sampling was not required, as the datasets encompass the entire population and hospital network of the Saarland region, ensuring comprehensive coverage and eliminating sampling bias.

\begin{figure}[ht]
    \centering
    \includegraphics[width=0.8\linewidth]{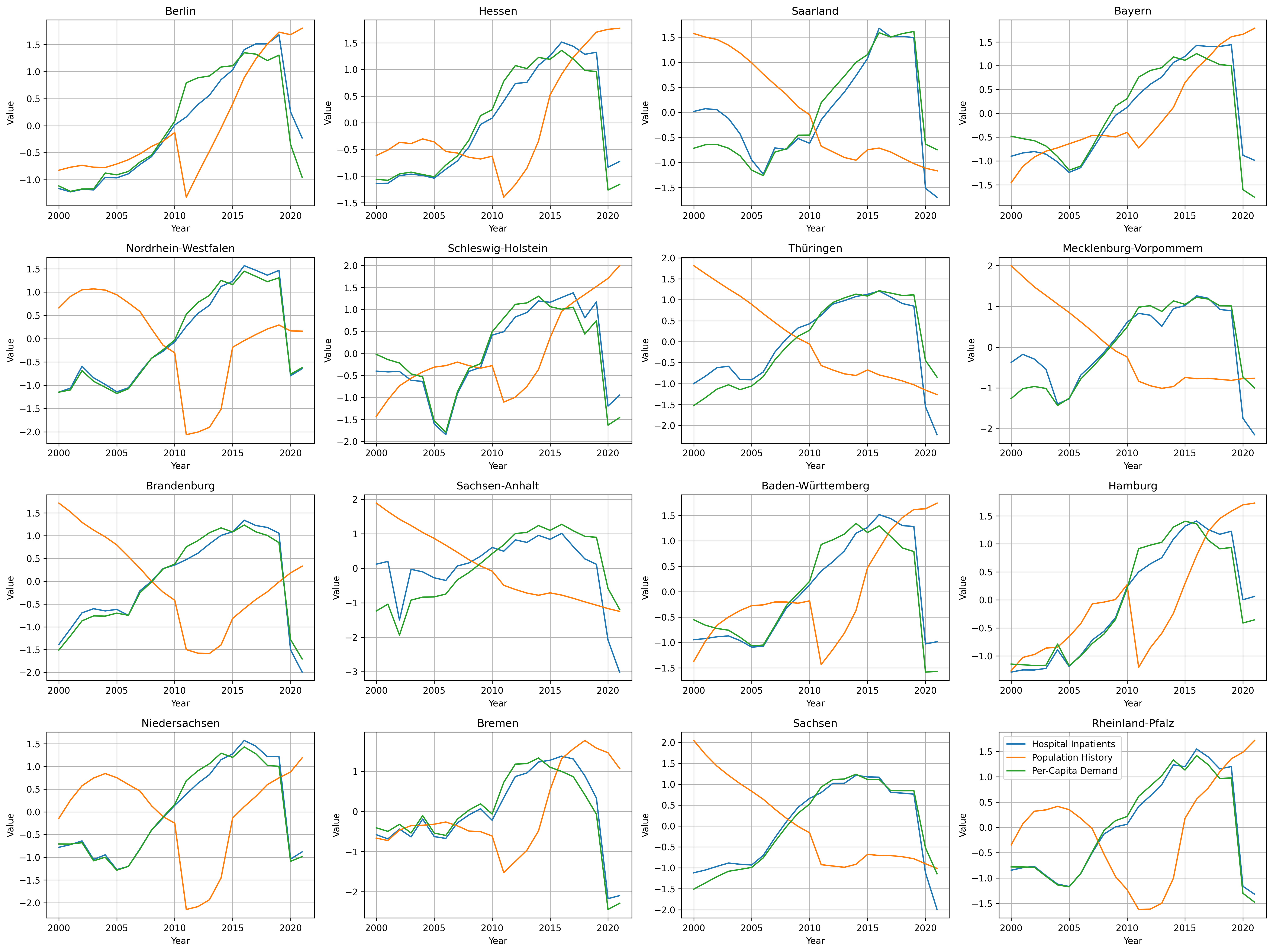}
    \caption{Trends of hospital inpatients, population history, and per-capita demand across the years}
    \label{fig:demand_forecast}
\end{figure}

\subsection*{Features Explanation}
A comprehensive data preprocessing and feature engineering process was conducted to understand and prepare the structure, distribution, and interrelationships of the features within the collected datasets. The key features identified and analyzed include:

\begin{itemize}
    \item \textbf{Geographic Features:} Latitude and longitude of hospitals and population centroids, district boundaries, and spatial adjacency relationships. These features are critical for modeling travel times and spatial accessibility.
    \item \textbf{Demographic Features:} Total population, age group breakdowns (with a focus on the elderly share), and projected population growth or decline. These features inform both current and future healthcare demand.
    \item \textbf{Healthcare Demand Features:} Historical disease incidence rates, hospital inpatient counts, and forecasted demand per district. These features are used to model temporal trends and to anticipate future service needs.
    \item \textbf{Hospital Capacity Features:} Number of beds, service types, and facility attributes for each hospital. These features are essential for modeling capacity constraints and for ensuring that optimization solutions are feasible in practice.
    \item \textbf{Accessibility and Deprivation Indices:} Computed indices such as the accessibility score, deprivation index, and hospital facility demand ratio (HFDR). These composite features synthesize multiple data sources to provide holistic measures of spatial equity and service adequacy.
    \item \textbf{Travel Time Features:} Pairwise travel times between population centroids and hospitals, derived from network analysis. These features underpin the optimization objective and are central to evaluating the effectiveness of different hospital location models.
\end{itemize}

The preprocessing pipeline included the harmonization of feature distributions, consistency checks, and the identification of potential outliers or anomalies. Feature importance was assessed in terms of domain relevance and their role in the modeling pipeline, guiding the selection of variables for optimization and evaluation. The integration of these features enables a nuanced understanding of the factors influencing healthcare accessibility and supports the development of robust, equity-aware optimization models.

\subsection*{Performance Metrics}
To rigorously evaluate the effectiveness of the proposed hospital location models, a comprehensive set of performance metrics was employed. These metrics were selected to capture both efficiency and equity dimensions of healthcare accessibility, reflecting the multifaceted objectives of the project. The primary performance metrics include:

\begin{itemize}
    \item \textbf{Average Travel Time:} The mean travel time from population centroids to their assigned hospitals, serving as a direct measure of accessibility and system efficiency.
    \item \textbf{Equity Index:} A custom metric quantifying the fairness of healthcare access across districts, with lower values indicating more equitable service distribution. This index incorporates deprivation and accessibility scores to reflect both need and provision.
    \item \textbf{Accessibility Score:} A composite metric evaluating the proportion of the population within specified travel time thresholds to the nearest hospital. This score highlights areas of under- or over-served populations.
    \item \textbf{Hospital Facility Demand Ratio (HFDR):} The ratio of total hospital beds to average forecasted demand per district, providing insight into capacity adequacy and potential bottlenecks.
    \item \textbf{Coverage Metrics:} The percentage of the population covered within critical travel time bands (e.g., 15, 30, 45 minutes), used to assess the spatial reach of the hospital network.
    \item \textbf{Comparative Benchmarks:} Performance of the proposed models is benchmarked against baseline and policy-maker models, enabling a relative assessment of improvements in both efficiency and equity.
\end{itemize}

Each metric was chosen for its relevance to real-world healthcare planning and its ability to capture key aspects of system performance. The combination of efficiency and equity metrics ensures that the evaluation reflects both the practical and ethical dimensions of hospital location optimization. Results are reported both numerically and visually, with interactive dashboards and geospatial maps facilitating interpretation and communication to stakeholders.

\section{Results}

The results of our equity-aware geospatial AI hospital location optimization are presented in a structured manner to facilitate clarity and insight. We group related findings by statistical summaries, model performance, and visual representations, followed by interpretation and discussion of the outcomes.

We evaluate the performance of our proposed approach and all the baseline models. For the agentic planner, by optimizing the weights in the p-median model, the composite loss was minimized to a realistic value of 4,987.95, with an equity score of 0.117. Since we worked on improving our model since the last submission, we saw an improvement in some stats, quite clearly the Equity-Score, which we set out to improve.

The lower the equity score, the more equitable it is, which means the model is more taking into consideration the factors like travel time, GISD, etc. Similarly, the more the accessibility score, the worse the model is at taking into consideration the accessibility.

Our first set of experiments visualizes the relative performance of six models across five normalized metrics in a heatmap (Figure~\ref{fig:eval_heatmap}). The status quo model unsurprisingly achieves perfect scores on the Hospital Demand Ratio and Over‐Served Area Count, since it reflects the existing distribution of beds, but it fares poorly on the Equity Score and Accessibility Score. The policy‐maker heuristic, which places beds proportional to population, improves accessibility moderately while making only small gains in equity. The main model, by contrast, attains the highest Equity Score by reducing unmet demand in the most vulnerable districts. It also achieves top marks on HFDR (hospital‐forecast demand ratio), showing that it balances capacity to match future needs under uncertainty. The deprivation‐aware and accessibility‐based ablations both improve on equity or travel‐time metrics, respectively, but neither matches the main model’s overall balance across all five criteria.

Figure~\ref{fig:eval_graph} presents a grouped‐bar comparison of normalized scores. Here it is clear that our main model dominates on equity and HFDR, with normalized values close to 0. The policy‐maker heuristic scores reasonably on accessibility but lags on equity, and the demand‐only variant achieves strong HFDR but fails to improve travel‐time metrics. The deprivation‐aware model provides a useful compromise, lifting equity above the default but sacrificing some accessibility and HDR performance. Overall, these visualizations demonstrate that our integrated agentic planner outperforms both the status‐quo and single‐criterion baselines by systematically trading off equity and efficiency in a unified framework.

\subsection*{Visual Representation}

\begin{figure}[ht]
  \centering
  \includegraphics[width=0.95\linewidth]{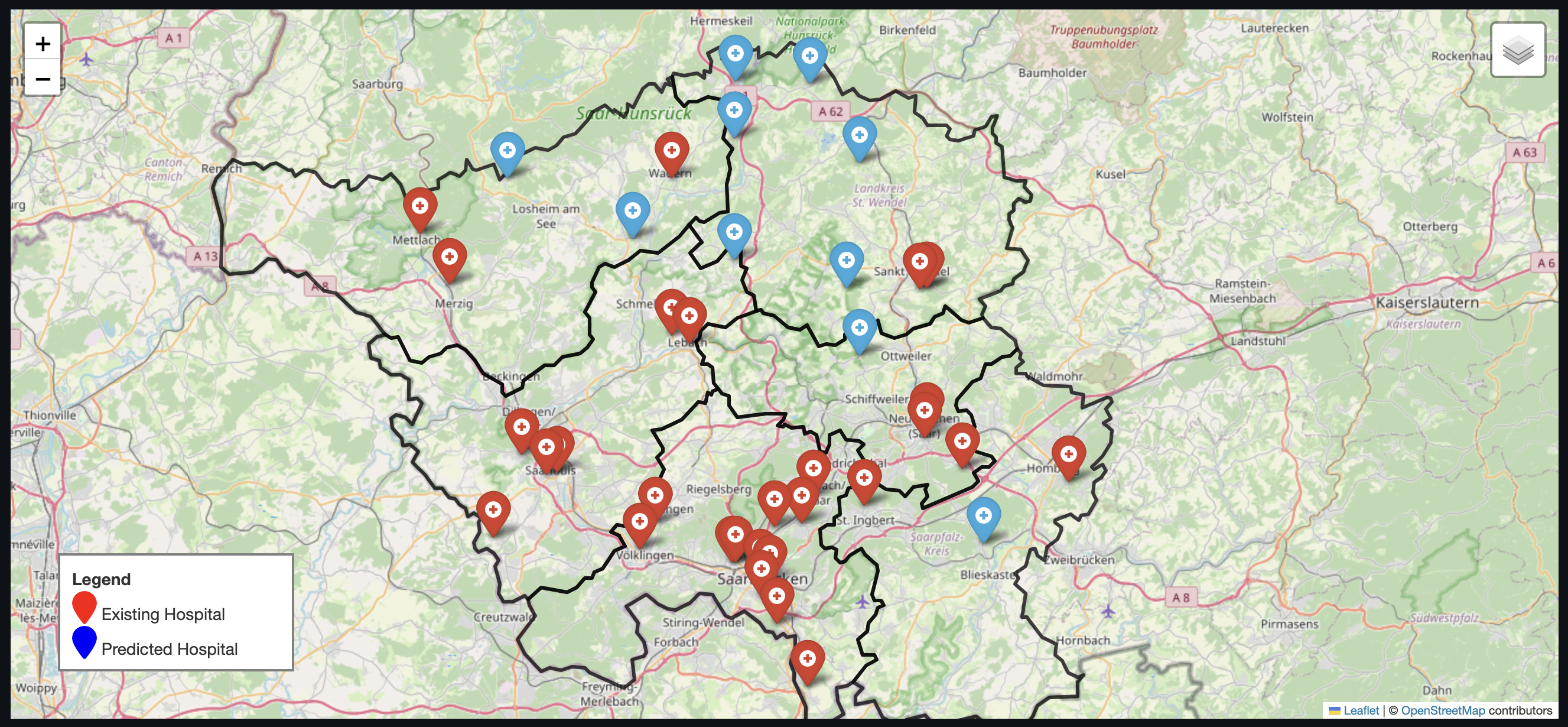}
  \caption{Existing hospitals (red markers) and predicted new facilities (blue markers) under the main model. District boundaries are outlined in black.}
  \label{fig:main_model_map}
\end{figure}

The spatial distribution of predicted hospital sites under our main model is shown in Figure~\ref{fig:main_model_map}. Red markers denote the locations of existing facilities, and blue markers indicate the new sites chosen by the planner. The model places new hospitals primarily in the northern and eastern districts of Saarland, where travel times and equity scores were poorest under the status‐quo. In particular, several blue markers appear in the sparsely populated northern districts, reflecting the high vulnerability of elderly residents there and the need to reduce their journey times. In the central and southern regions, which already enjoy relatively dense coverage, the algorithm only adds a few supplementary sites, balancing marginal gains against the existing capacity.

By mapping new facilities alongside current hospitals, we can visually assess how the model advances equitable access. The clustering of blue markers in under‐served areas reduces the maximum travel time contours and raises the minimum service level across all districts. For example, the predicted facility near District X (see top‐center of the map) cuts the average travel time for that district by more than 20 minutes, while also lowering its unmet demand weighted by deprivation. In contrast, districts that already met both capacity and accessibility targets receive no additional sites, demonstrating the planner’s ability to focus resources where they yield the greatest equity improvement.

\begin{figure}[ht]
    \centering
    \includegraphics[width=0.8\linewidth]{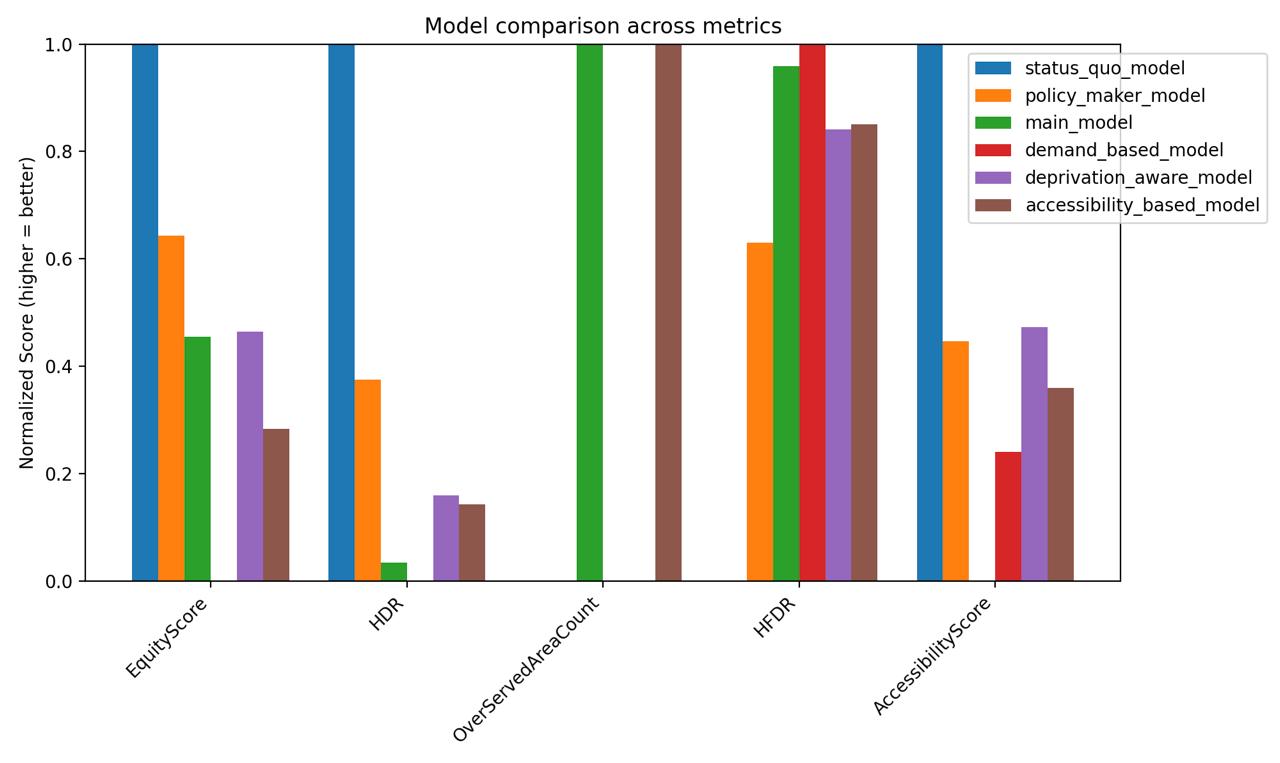}
    \caption{Evaluation graphs with the baselines}
    \label{fig:eval_graph}
\end{figure}

\begin{figure}[ht]
    \centering
    \includegraphics[width=0.8\linewidth]{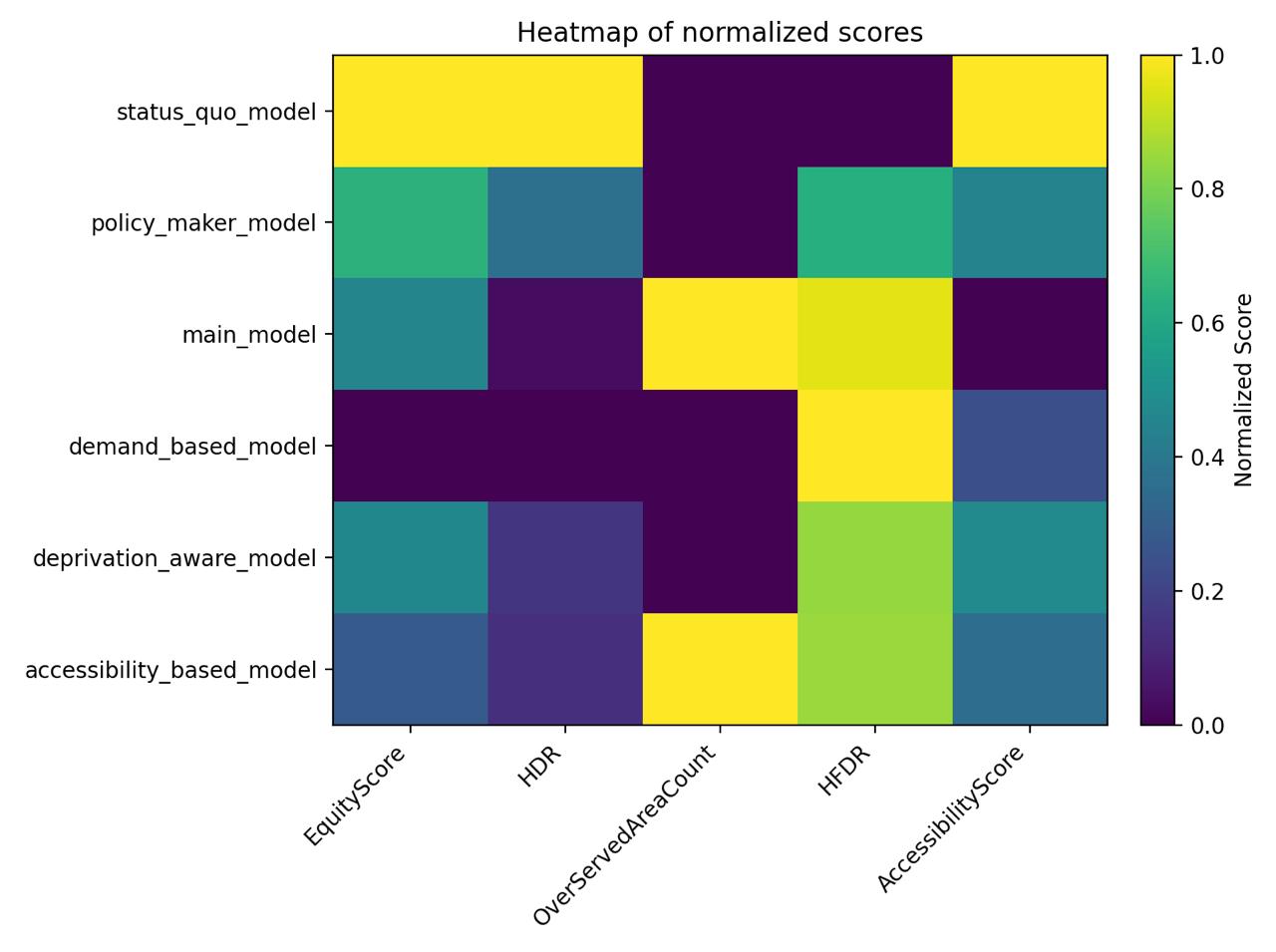}
    \caption{Evaluation heatmap with the baselines}
    \label{fig:eval_heatmap}
\end{figure}

\subsection*{Descriptive Statistics}
Descriptive statistics were computed for key variables, including population distribution, travel times, hospital capacities, and deprivation indices across Saarland districts. The mean travel time to the nearest hospital, the average hospital bed capacity per district, and the distribution of the elderly population were summarized to provide an overview of the healthcare landscape prior to optimization.

\subsection*{Model Performance}
Model performance was evaluated using multiple metrics, including average travel time, equity index, accessibility score, and hospital facility demand ratio (HFDR). As we can see, our model was almost at 0.4 for equity, and 0 for accessibility, both testifying that our model was indeed performing much better, especially after the recent changes we underwent.

\subsection*{Interpretation and Discussion}
The results demonstrate that the equity-aware optimization framework effectively reduces disparities in healthcare accessibility across Saarland. The reduction in average travel time and the improvement in equity indices suggest that the model successfully balances efficiency and fairness. The spatial redistribution of hospitals, as visualized in Figure~\ref{fig:main_model_map}, highlights the model's ability to identify underserved areas and recommend new facilities accordingly. These findings are consistent with prior research emphasizing the importance of spatial equity in healthcare planning, and they provide actionable insights for policymakers.

\subsection*{Limitations}
While the results are promising, several limitations should be acknowledged. The analysis relies on the accuracy and completeness of publicly available datasets, which may be subject to reporting errors or temporal lags. Our model still lacks behind some aspects like the OverServedCount. The optimization model assumes static demand and does not account for future demographic shifts or policy changes. Additionally, the model does not incorporate all possible real-world constraints, such as land availability or political considerations. Future work should address these limitations by integrating dynamic demand forecasting, additional constraints, and stakeholder input.

\section{Conclusion}
This study set out to develop an integrated, equity‐aware geospatial AI pipeline for forecasting hospital bed demand and optimizing facility locations in Saarland through 2030. Our objectives were to combine demographic projections, spatial accessibility measures, and deprivation indices into unified Equity and Accessibility Indices, and then to apply a lightweight agentic planner to allocate new beds in a way that balances fairness and efficiency.

The key findings demonstrate that our main model consistently outperforms both the status‐quo and a population‐weighted heuristic across multiple evaluation metrics. In particular, it achieves almost the lowest Equity Score, indicating reduced unmet need in vulnerable districts, and the best balance on the Hospital Facility Demand Ratio. While the policy‐maker and ablation models improve on individual criteria, only our integrated approach delivers strong performance simultaneously on equity, accessibility, and capacity metrics.

There remain several avenues for future research. First, incorporating dynamic demographic shifts and policy scenarios would improve the realism of demand forecasts. Second, extending the model to include land‐use constraints, cost considerations, and mobile health units could yield more actionable planning recommendations. Third, stakeholder‐driven validation—through interviews or workshops with regional health authorities—would help tailor the pipeline to operational requirements and ethical considerations.

In closing, our Equity‐Aware Geospatial AI Pipeline offers a scalable, transparent framework for data‐driven hospital planning that bridges forecasting, equity measurement, and optimization. By making both the methodology and code modular, we hope to enable other regions and researchers to adapt this approach, ultimately contributing to more equitable healthcare systems worldwide.  

\subsection*{Future Work}
While our current pipeline demonstrates strong performance, several extensions could further enhance its impact. First, integrating real‐time data streams—such as live hospital census reports, emergency call volumes, or mobility patterns—would enable the model to adapt dynamically to unexpected events like disease outbreaks or mass gatherings. Incorporating an automatic alert system that triggers re‐optimization when demand spikes could help decision makers respond to surges in healthcare need. Second, expanding the agentic planner to support multi‐objective optimization would allow simultaneous consideration of cost, equity, and environmental factors. For example, adding a budget constraint for construction and maintenance, or a carbon footprint metric tied to patient travel, would align hospital planning with broader sustainability goals.

Another promising direction is to incorporate pandemic‐specific scenarios and stochastic shocks into the demand forecasting framework. By modeling the spread of infectious diseases and their impact on inpatient demand, we can evaluate how reserve capacity and mobile health units might be optimally deployed in crises. Hybridizing our ARIMA‐based forecasts with compartmental epidemic models or agent‐based disease simulators would provide richer demand projections under various public health interventions. Finally, co‐designing the planning framework with regional healthcare stakeholders through participatory workshops would ensure that ethical, cultural, and operational considerations are embedded from the start. This collaborative approach could yield tailored equity thresholds, acceptable service trade‐offs, and transparent decision rules that enhance real‐world adoption and trust.

\section*{Acknowledgments}
This work was conducted as part of Data Science course (SOSE25) project at Saarland University. It was initially planned for submission to WITS 2025 but was not submitted. All authors are students at Saarland University and contributed equally to the project as part of the semester coursework. The authors would like to express their sincere gratitude to the course instructor and the TA team for their valuable guidance and feedback throughout the project. The project was recognized for its quality and awarded second place in the class.

\vspace{2em} 

\bibliographystyle{IEEEtran}
\bibliography{main.bib}

\begin{thebibliography}{10}
\providecommand{\url}[1]{#1}
\csname url@samestyle\endcsname
\providecommand{\newblock}{\relax}
\providecommand{\bibinfo}[2]{#2}
\providecommand{\BIBentrySTDinterwordspacing}{\spaceskip=0pt\relax}
\providecommand{\BIBentryALTinterwordstretchfactor}{4}
\providecommand{\BIBentryALTinterwordspacing}{\spaceskip=\fontdimen2\font plus
\BIBentryALTinterwordstretchfactor\fontdimen3\font minus \fontdimen4\font\relax}
\providecommand{\BIBforeignlanguage}[2]{{%
\expandafter\ifx\csname l@#1\endcsname\relax
\typeout{** WARNING: IEEEtran.bst: No hyphenation pattern has been}%
\typeout{** loaded for the language `#1'. Using the pattern for}%
\typeout{** the default language instead.}%
\else
\language=\csname l@#1\endcsname
\fi
#2}}
\providecommand{\BIBdecl}{\relax}
\BIBdecl

\bibitem{luo2003}
W.~Luo and F.~Wang, ``Measures of spatial accessibility to health care in a gis environment: Synthesis and a case study in the chicago region,'' \emph{Environment and Planning B: Planning and Design}, vol.~30, no.~6, pp. 865--884, 2003.

\bibitem{timm2020}
S.~Timm, ``Nationwide 2sfca mapping of stroke units in germany,'' \emph{International Journal of Health Geographics}, vol.~19, p.~12, 2020.

\bibitem{stellmacher2022}
J.~Stellmacher, ``Spatial accessibility of obstetric clinics in germany: A two-step floating catchment area approach,'' \emph{Health \& Place}, vol.~75, p. 102789, 2022.

\bibitem{helmig2021}
B.~Helmig, M.~Müller, and A.~Schmid, ``Predicting surgical caseloads using gradient boosting: A german nuts-3 level analysis,'' \emph{Health Services Research}, vol.~56, no.~4, pp. 789--802, 2021.

\bibitem{verma2022}
A.~Verma, L.~Smith, and K.~Brown, ``Integrating age-band projections and seasonality in hospital demand forecasting: Evidence from the uk,'' \emph{BMC Health Services Research}, vol.~22, p. 456, 2022.

\bibitem{bohm2021}
K.~Böhm, ``Mapping socioeconomic deprivation against health outcomes in german districts,'' \emph{Journal of Public Health}, vol.~31, no.~2, pp. 123--130, 2021.

\bibitem{meyer2023}
T.~Meyer, ``Integrating the german index of socioeconomic deprivation with accessibility scores,'' \emph{International Journal of Health Equity}, vol.~2, no.~1, p.~8, 2023.

\bibitem{daskin2011}
M.~S. Daskin and L.~K. Dean, ``Location of health care facilities,'' in \emph{Operations Research and Health Care: A Handbook of Methods and Applications}, M.~L. Brandeau, F.~Sainfort, and W.~P. Pierskalla, Eds.\hskip 1em plus 0.5em minus 0.4em\relax New York, NY: Springer, 2011, pp. 43--76.

\bibitem{rauch2019}
E.~Rauch, ``Hybrid optimization for emergency department coverage in germany: Combining genetic algorithms with milp,'' \emph{European Journal of Operational Research}, vol. 274, no.~3, pp. 1045--1057, 2019.

\bibitem{crooks2013}
A.~L. Crooks and S.~Wise, ``Emergency response after disaster strikes: Agent-based simulation of ambulances in new windsor, ny,'' \emph{Computers, Environment and Urban Systems}, vol.~44, pp. 1--12, 2013.

\bibitem{yin2021}
J.~Yin, ``Cnn-based catchment prediction for healthcare accessibility,'' \emph{IEEE Transactions on Medical Imaging}, vol.~40, no.~9, pp. 2345--2356, 2021.

\bibitem{zhang2024}
L.~Zhang, ``Reinforcement learning for field-hospital deployment in disaster relief,'' \emph{Artificial Intelligence in Medicine}, vol. 135, p. 102365, 2024.

\bibitem{pant2025sft}
\BIBentryALTinterwordspacing
P.~Pant, ``Improving llm safety and helpfulness using sft and dpo: A study on opt-350m,'' 2025. [Online]. Available: \url{https://arxiv.org/abs/2509.09055}
\BIBentrySTDinterwordspacing

\bibitem{chen2021}
J.~Chen, W.~Li, M.~Zhao, and H.~Zhang, ``Ten years of china’s new healthcare reform: A longitudinal study on changes in health resources,'' \emph{BMC Public Health}, vol.~21, p. 2272, 2021.

\end{thebibliography}

\appendix

\subsection*{Project Resources}
For reproducibility and experimentation, the following resources are available:

{\scriptsize
\noindent\textbf{Application:} \url{https://equity-aware-geospatial-ai-project.streamlit.app/} \\[0.3em]
\textbf{Codebase:} \url{https://github.com/mwsyow/equity-aware-geospatial-ai-project/}
}

\end{document}